\documentclass[11pt,a4paper]{article}
\usepackage[hyperref]{acl2019}
\usepackage{times}
\usepackage{latexsym}
\usepackage{graphicx}
\usepackage{url}
\usepackage{amsmath}
\usepackage{subfigure}
\usepackage{amssymb}
\usepackage{multirow}
\usepackage{color}
\usepackage{colortbl}
\usepackage{CJKutf8}
\usepackage{booktabs}
\usepackage{array} 
\usepackage{mwe}
\usepackage{titlesec}
\usepackage{enumitem}

\newcolumntype{R}[1]{>{\raggedleft\arraybackslash}p{#1}}
\newcolumntype{L}[1]{>{\raggedright\arraybackslash}p{#1}}

\aclfinalcopy 

\def\LM{{\mathcal L}}
\def\MM{{\mathcal M}}

\def\UM{{\mathcal U}}
\def\VM{{\mathcal V}}

\def\c{{\bf c}}

\def\0{{\bf 0}}
\def\1{{\bf 1}}

\title{ZEN: Pre-training Chinese Text Encoder\\ Enhanced by N-gram Representations}

\author{Shizhe Diao$^\heartsuit$\thanks{~~Work done during the internship at Sinovation Ventures.}~, ~ Jiaxin Bai$^{\heartsuit*}$, ~ Yan Song$^\spadesuit$, ~ Tong Zhang$^\heartsuit$, ~ Yonggang Wang$^\spadesuit$\\
  $^\heartsuit$The Hong Kong University of Science and Technology\\
  \texttt{\{sdiaoaa, jbai, tongzhang\}@ust.hk}\\
  $^\spadesuit$Sinovation Ventures\\
  \texttt{\{songyan, wangyonggang\}@chuangxin.com}
}

\date{}

\begin{document}
\begin{CJK}{UTF8}{gbsn}
\maketitle
\begin{abstract}
%
The pre-training of text encoders normally processes text as a sequence of tokens corresponding to small text units, such as word pieces in English and characters in Chinese. It omits information carried by larger text granularity, and thus
%
the encoders cannot easily adapt to certain combinations of characters. This leads to a loss of important semantic information, which is especially problematic for Chinese because the language does not have explicit word boundaries.
%
%
In this paper, we propose ZEN, a BERT-based Chinese (\textbf{Z}) text encoder \textbf{E}nhanced by \textbf{N}-gram representations,
where different combinations of characters are considered during  training.
%
As a result, potential word or phrase boundaries are explicitly pre-trained and fine-tuned with the character encoder (BERT).
Therefore ZEN incorporates the comprehensive information of both the character sequence and words or phrases it contains.
Experimental results illustrated the effectiveness of ZEN on a series of Chinese NLP tasks.
We show that ZEN, using less resource than other published encoders, can achieve state-of-the-art performance on most tasks.
Moreover, it is shown that reasonable performance can be obtained when ZEN is trained on a small corpus,
which is important for applying pre-training techniques to scenarios with limited data.
The code and pre-trained models of ZEN are available at \url{https://github.com/sinovation/ZEN}.

\end{abstract}

\section{Introduction}

Pre-trained text encoders \cite{peters2018deep, devlin2018bert, radford2018improving, radford2019language, yang2019xlnet} have drawn much attention in natural language processing (NLP),
because state-of-the-art performance can be obtained for many NLP tasks using such encoders.
In general, these encoders are implemented by training a deep neural model on large unlabeled corpora. 
%
%
%
Although the use of big data brings success to these pre-trained encoders, it is still unclear whether existing encoders have effectively leveraged all useful information in the corpus.
Normally, the pre-training procedures are designed to learn on tokens corresponding to small units of texts (e.g., word pieces for English, characters for Chinese) for efficiency and simplicity. However, some important information carried by larger text units may be lost for certain languages when we use a standard encoder, such as BERT.
%
For example, in Chinese, text semantics are greatly affected by recognizing valid n-grams\footnote{Herein `valid' regards to that an n-gram is a proper chunk or phrase that is frequently used in the running text.}. This means a pre-trained encoder can potentially be improved by incorporating such boundary information of important n-grams.
%

Recently, there are studies adapting BERT for Chinese with word information, yet they are limited in maintaining the original BERT structure, augmented with learning from weakly supervised word information or requiring external knowledge.
%
%
As an example, a representative study in \citet{cui2019pre} proposed to use the whole-word masking strategy to mitigate the limitation of word information. 
They used an existing segmenter to produce possible words in the input sentences, and then train a standard BERT on the segmented texts by masking whole words.
\citet{sun2019ernie} proposed to perform both entity-level and phrase-level masking to learn knowledge and information from the pre-training corpus.
However, their approaches are limited in the following senses.
First, both methods rely on the word masking strategy so that the encoder can only be trained with existing word and phrase information. Second,
similar to the original BERT, the masking strategy results in the mis-match of pretraining and fine-tuning, i.e.,
no word/phrase information is retained when the encoders are applied to downstream prediction tasks. 
Third, incorrect word segmentation or entity recognition results cause errors propagated to the pre-training process and thus may negatively affected the generalization capability of the encoder.

In this paper, we propose ZEN, a Chinese (\textbf{Z}) text encoder \textbf{E}nhanced by representing \textbf{N}-grams,
which provides an alternative way to improve character based encoders (e.g., BERT) by using larger text granularity.
%
To train our model, one uses an n-gram lexicon from any possible sources such as pre-defined dictionaries and n-gram lists extracted via unsupervised approaches.
Such lexicon is then mapped to training texts, and is used to highlight possible combinations of characters that indicate likely salient contents during the training process. Our model then integrate the representations of these n-gram contexts with the character encoder. 
Similarly, the fine-tune process on any task-specific dataset further enhances ZEN with such n-gram representations. An important feature of our method is that while the model explicitly takes advantage of n-gram information, the model only outputs character-level encodings that is consistent with BERT. Therefore downstream tasks are not affected.
ZEN extends the original BERT model and incorporate learning from large granular text explicitly into the model, which is different (and complementary) from previous methods that relied on weak supervision such as whole-word masking.\footnote{Although the character encoder may still use masking as a learning objective, the encoded n-grams are explicitly leveraged in our model.}
%
Our experiments follow the standard procedure, i.e., training ZEN on the Chinese Wikipedia dump and fine-tune it on several Chinese downstream NLP tasks.
Experiment results demonstrate its validity and effectiveness where state-of-the-art performance is achieved on many tasks using the n-grams automatically learned from the training data other than external or prior knowledge.
In particular, our method outperforms some existing encoders trained on much larger corpora on these tasks.

\begin{figure*}[t]
\includegraphics[scale=0.42, trim=0 10 0 30]{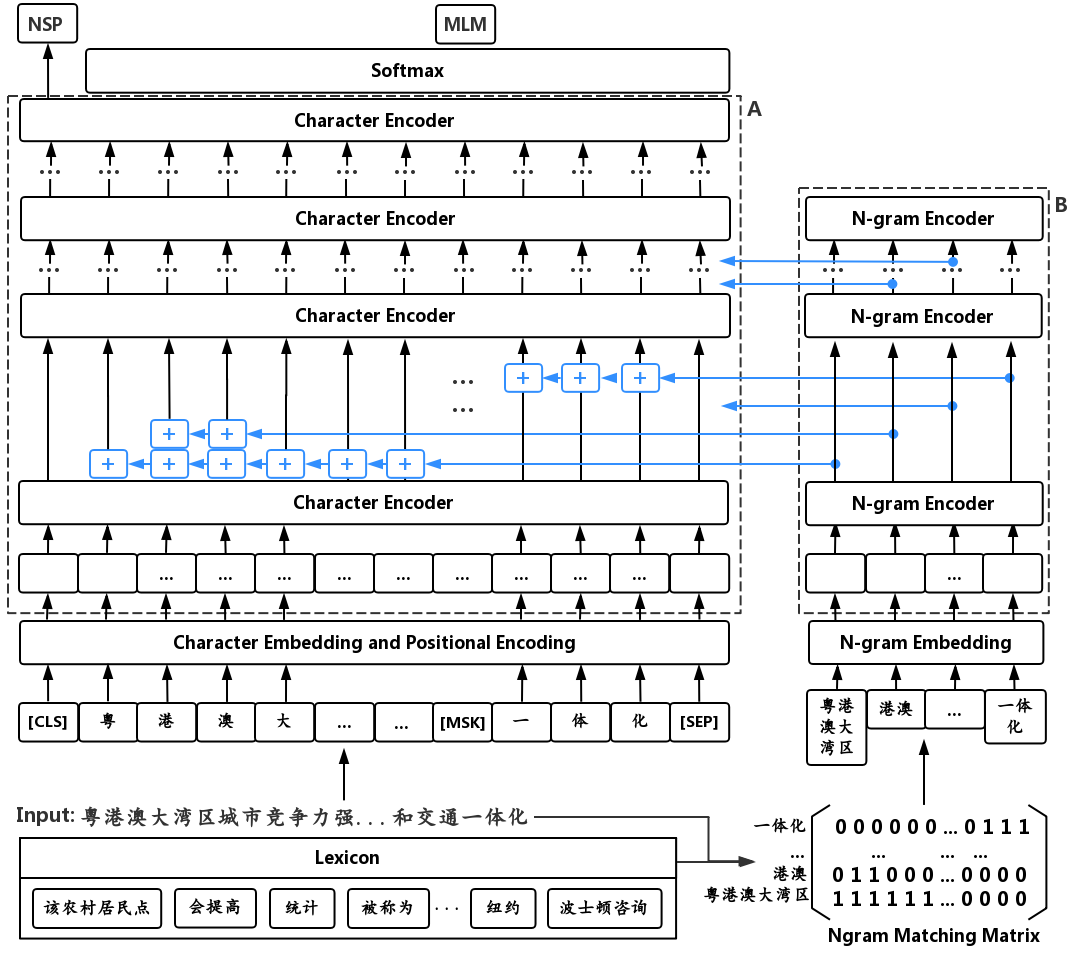}
\caption{
The overall architecture of ZEN, where the area marked by
dashed box `A' presents the character encoder (BERT, in Transformer structure);
and the area marked by dashed box `B' is the n-gram encoder.
[NSP] and [MLM] refer to two BERT objectives: next sentence prediction and masked language model, respectively. [MSK] is the masked token.
The incorporation of n-grams into the character encoder is illustrated by the addition operation presented in blue color.
The bottom part presents n-gram extraction and preparation for the given input instance.
%
}
\label{modelgraph}
\vskip -0.5em
\end{figure*}

\section{ZEN}
The overall architecture of ZEN is shown in Figure \ref{modelgraph}, where
the backbone model (character encoder) is BERT\footnote{The two terms, `BERT' and `character encoder' are used interchangeably in this paper.} \cite{devlin2018bert}, enhanced by n-gram information represented by a multi-layer encoder.
Since the basis of BERT is well explained in previous studies \citep{devlin2018bert,yu2019adapting},
in this paper, we focus on the details of ZEN, by explaining how n-grams are processed and incorporated into the character encoder.


\subsection{N-gram Extraction}

Pre-training ZEN requires n-gram extraction in the first place before training starts,
where two different steps are performed.
The first one is to prepare an n-gram lexicon, $\LM$, 
from which one can use any unsupervised method to extract n-grams for later processing.
The second step of n-gram extraction is performed during pre-training, where some n-grams in $\LM$ are selected according to each training instance $\c = (c_1, c_2, ..., c_i, ..., c_{k_c})$ with $k_c$ characters.
Once these n-grams are extracted, we use an n-gram matching matrix, $\MM$, to record the positions of the extracted n-grams in each training instance.
$\MM$ is thus an $k_c \times k_n$ matrix, where each element is represented by
$$ m_{ij}=\left\{
\begin{aligned}
1 &  &{c_i \in n_j}\\
0 &  &{c_i  \not\in n_j }
\end{aligned}
\right. ,
$$
where $k_n$ is the number of extracted n-grams from $\c$, and  $n_j$ denotes the $j$-th extracted n-gram.
A sample $\MM$ with respect to an input text is shown in the bottom part of Figure \ref{modelgraph}.



\subsection{Encoding N-grams}

As shown in the right part of Figure \ref{modelgraph} (dashed box marked as `B'),
ZEN requires a multi-layer encoder
to represent all n-grams,
whose information are thus encoded in different levels
matching the correspondent layers in BERT.
We adopt Transformer \cite{vaswani2017attention} as the encoder,
which is a multi-layer encoder that can model the interactions among all n-grams through their representations in each layer.
This modeling power is of high importance for ZEN because for certain context, salient n-grams are more useful than random others,
and such salient n-grams are expected to be emphasized in pre-training.
This effect can be achieved by
multi-head self-attention (MhA) mechanism in Transformer \cite{clark2019does}.
In detail, the transformer for n-grams is the same as its original version for sequence modeling, except that it does not encode n-gram positions because all n-grams are treated equally without a sequential order.
%
%
Formally, denote the $j$-th n-gram in layer $l$  by $\mu^{(l)}_j$,
%
the n-gram encoder represents each of them 
by MhA via
\begin{equation} \label{equation_transformer}
\mu^{(l+1)}_j = MhA(Q=\mu^{(l)}_j, K=V=\UM^{(l)})
\end{equation}
%
where 
$\mu^{(l)}_j$ is used as the query ($Q$) vector to calculate the attentions over all other input n-grams from the same layer,
and $\UM^{(l)}$ refers to the matrix that stacks all n-gram representations in the layer $l$ that servers as the key ($K$) and value ($V$) in MhA.
This encoding process
is repeated layer-by-layer along with the character encoder.

\begin{table*}[t]
\begin{center}
\small
\begin{sc}
\begin{tabular}{p{1.2cm}|R{0.6cm}R{0.6cm}R{0.6cm}R{0.6cm}R{0.6cm}R{0.6cm}R{0.6cm}R{0.6cm}R{0.6cm}R{0.6cm}R{0.6cm}R{0.6cm}R{0.6cm}R{0.6cm}}
\toprule
Task & \multicolumn{2}{c}{CWS} & \multicolumn{2}{c}{POS} & \multicolumn{2}{c}{NER} & \multicolumn{2}{c}{DC} & \multicolumn{2}{c}{SA} & \multicolumn{2}{c}{SPM} & \multicolumn{2}{c}{NLI} \\
\midrule
Dataset & \multicolumn{2}{c}{MSR} & \multicolumn{2}{c}{CTB5} & \multicolumn{2}{c}{MSRA} & \multicolumn{2}{c}{News} & \multicolumn{2}{c}{CSC} & \multicolumn{2}{c}{LCQMC} & \multicolumn{2}{c}{XNLI}\\
\cline{2-15}
 & S\# & C\# & S\# & C\# & S\# & C\# & D\# & C\# & D\# & C\# & SP\# & C\# & SP\# & C\#\\
\midrule
Train & 87K & ~4M & 18K & 720K & 45K & 2M & 50K & 41M & 10K & 927K & 239K & 5M & 393K & 23M \\
Dev  & - & - & 350 & 10K & - & - & 5K & 4M & 1K & 115K & 9K & 209K & 3K & 136K \\
Test & 4K & 173K & 348 & 13K & 3K & 153K & 10K & 9M & 1K & 114K & 13K & 233K &  3K & 273K \\
\bottomrule
\end{tabular}
\end{sc}
\caption{The statistics of task datasets used in our experiments. 
S\#, C\#, D\# and SP\# refer to numbers of sentences, characters, documents and sentence pairs, respectively.}
\label{tab:StatisticsofDatasets}
\end{center}
\vskip -1em
\end{table*}

\subsection{Representing N-grams in Pre-training}

%
%
With the n-gram encoder, ZEN combine the representations of each character and its associated n-grams
to train the backbone model, as shown in the left upper part of Figure \ref{modelgraph} (dashed box marked as `A').
In detail, let $\upsilon_i^{(l)}$ and $\mu_{i,k}^{(l)}$ represent embeddings for the $i$-th character and the $k$-th n-gram associated to this character at layer $l$, 
the enhanced representation for this character is computed by
\begin{equation}\label{equation_character}
\setlength\belowdisplayskip{1pt plus 3pt minus 5pt}
\upsilon_i^{(l)*} =\upsilon_i^{(l)} + \sum_k {\mu_{i,k}^{(l)}}
\end{equation}
where $\upsilon_i^{(l)*}$ is the resulting embedding sent to the next layer.
Herein $+$ and $\sum$ refer to the element-wise addition operation.
Therefore, $\upsilon_i^{(l)*} = \upsilon_i^{(l)}$ when no n-gram covers this character.
For the entire layer $l$, this enhancement can be formulated by
\begin{equation} \label{equation_character_matrix}
\VM^{(l)*} = \VM^{(l)} + \MM\times\UM^{(l)}
\end{equation}
where $\VM^{(l)}$ is the embedding matrix for all characters,
and its combination with $\UM^{(l)}$ can be directly done through $\MM$.
This process is repeated for each layer in the backbone BERT excecept for the last one. The final output of all character embeddings from the last layer is sent to optimize BERT objectives, i.e., mask recovery and next sentence prediction.
Note that, since there is masking in BERT training, when a character is masked,
n-grams that cover this character are not considered.

\section{Experiment Settings}



\subsection{Tasks and Datasets}


For pre-training, following previous studies \cite{devlin2018bert,cui2019pre}, we use Chinese Wikipedia dump\footnote{https://dumps.wikimedia.org/zhwiki/} as the base corpus to learn different encoders including ZEN.
To clean the base corpus, we remove useless symbols and translate all traditional characters into simplified ones, and lowercase all English letters.
The resulted corpus contains 474M tokens and 23K unique characters.
For fine-tuning, we choose seven NLP tasks and their corresponding benchmark datasets in our experiments, many of them have been used in previous studies \citep{cui2019pre, sun2019ernie, sun2019ernie2}.
These tasks and datasets are described as follows.



\vspace{-0.15cm}
\begin{itemize}[leftmargin=10pt]
    \itemsep-0.15em
\item{\textbf{Chinese word segmentation (CWS)}}: MSR dataset from SIGHAN2005 Chinese word segmentation Bakeoff \citep{emerson2005second}.

\item{\textbf{Part-of-speech (POS) tagging}}: CTB5 \citep{Xue2005The} dataset with standard splits.

\item{\textbf{Named entity recognition (NER)}}: MSRA dataset from
international Chinese language processing Bakeoff 2006\footnote{http://sighan.cs.uchicago.edu/bakeoff2006/}.
\item{\textbf{Document classification (DC)}}: 
THUCNews (News) dataset
\citep{sun2016thuctc} from Sina news with 10 evenly distributed classes.
\item{\textbf{Sentiment analysis (SA)}}:
The ChnSentiCorp\footnote{https://github.com/pengming617/bert\_classification} (CSC) dataset with 12,000 documents from three domains, i.e., book, computer and hotel.
\item{\textbf{Sentence pair matching (SPM)}}: The LCQMC (a large-scale Chinese question matching corpus) proposed by \citet{liu2018lcqmc},
where each instance is a pair of two sentences with a label indicating whether their intent is matched.
\item{\textbf{Natural language inference (NLI)}}:
The Chinese part of the XNLI
\citep{conneau2018xnli}.
\end{itemize}
\vspace{-0.25cm}
The statistics of these datasets with respect to their splits are reported in Table \ref{tab:StatisticsofDatasets}.
For CWS, POS, we fine-tune and test according to their standard split of training and test sets.
For the other tasks, we follow the settings of \citet{cui2019pre} to process those datasets in our experiments.




\begin{table*}[t]
\begin{center}
\small
\begin{sc}
\begin{tabular}{p{2.6cm}R{0.65cm}R{0.65cm}R{0.65cm}R{0.65cm}R{0.65cm}R{0.65cm}R{0.65cm}R{0.65cm}R{0.65cm}R{0.65cm}R{0.65cm}R{0.65cm}}
\toprule
 & 
CWS & \multicolumn{2}{c}{POS} & NER & \multicolumn{2}{c}{DC} & \multicolumn{2}{c}{SA} & \multicolumn{2}{c}{SPM} & \multicolumn{2}{c}{NLI} \\

\cline{2-13}
& Test & Dev & Test & Test & Dev & Test & Dev & Test & Dev & Test & Dev & Test   \\ 
\midrule
BERT (R)      & 97.20 & 95.72 & 95.43 & 93.12 & 96.90 & 96.71 & 94.00 & 94.10 & 87.22 & 85.13 & 75.67 & 75.01 \\
BERT (P)      & 97.95 & 96.30 & 96.10 & 94.78 & 97.60 & 97.50 & 94.53 & 94.67 & 88.50 & 86.59 & 77.40 & 77.52 \\
\hline
BERT-wwm   & - & - & - & 95.10 & 97.60 & 97.60 & 94.50 & 95.00 & 89.20 & 86.80& 78.40 & 78.00\\
ERNIE 1.0   & - & - & - & 95.10 & 97.30 & 97.30 & 95.20 & 95.40 & 89.70 & 87.40 & 79.90 & 78.40 \\
ERNIE 2.0 (B) & - & - & - & - & - & - & \textbf{95.70} & 95.50 & \textbf{90.90} & 87.90 & 81.20 & \textbf{79.70} \\
NEZHA (B) & - & - & - & - & - & - & 94.74 & 95.17 & 89.98 & 87.41 & \textbf{81.37} & 79.32 \\
NEZHA-wwm (B) & - & - & - & - & - & - & 94.75 & 95.84 & 89.85 & 87.10 & 81.25 & 79.11 \\
\hline
\textit{ERNIE 2.0 (L)} & - & - & - & - & - & - & \textit{96.10} & \textit{95.80} & \textit{90.90} &
\textit{87.90} & \textit{82.60} & \textit{81.00} \\
\textit{NEZHA (L)} & - & - & - & - & - & - & \textit{95.92} & \textit{95.83} & \textit{90.18} & \textit{87.20} & \textit{81.53} & \textit{80.44} \\
\textit{NEZHA-wwm (L)} & - & - & - & - & - & - & \textit{95.75} & \textit{96.00} & \textit{90.87} & \textit{87.94} & \textit{82.21} & \textit{81.17} \\
\hline
ZEN (R)       & 97.89 & 96.12 & 95.82 & 93.24 & 97.20 & 96.87 & 94.87 & 94.42 & 88.10 & 85.27 & 77.11 & 77.03 \\ 
ZEN (P)       & \textbf{98.35} & \textbf{97.43} & \textbf{96.64} & \textbf{95.25} & \textbf{97.66} & \textbf{97.64} & 95.66 & \textbf{96.08} & 90.20 & \textbf{87.95} & 80.48 & 79.20  \\ 
\bottomrule
\end{tabular}
                \end{sc}
                \vspace{-0.3cm}
\end{center}
\caption{The overall performance of ZEN and the comparison against existing models on seven NLP tasks, where R denotes that pre-training starts from random initialization and P is that model parameters are initialized from Google's released Chinese BERT base model.
B and L refer to each backbone model uses BERT base or large model, respectively.
Since ZEN uses BERT base model, encoders using BERT large model and their performance are listed as references in italic fonts.
The bold numbers are the best results from all base models in each column.
}
\label{tab:OverallPerformance}
\vskip -0.5em
\end{table*}

\subsection{Implementation}
N-grams to build the lexicon $\LM$ are extracted from the same training corpus, i.e., Chinese Wikipedia dump, and prepared by sorting them (except for unigrams) according to their frequencies.
We try the cut-off threshold between 5 and 40 where all those n-grams with frequency lower than the threshold are not included in $\LM$.
As a result, the sizes of $\LM$ with respect to different threshold range from 179K to 64K n-grams in them.\footnote{Our main experiments are conducted on cut-off=15, resulting in 104K n-grams in the lexicon.}
The embeddings of the n-grams are randomly initialized.

For the backbone BERT in ZEN, we use the same structure as that in previous work \citep{devlin2018bert, sun2019ernie, cui2019pre}, i.e., 12 layers with 12 self-attention heads, 768 dimensions for hidden states and 512 for max input length, etc.
The pre-training tasks also employ the same masking strategy and next sentence prediction as in \citet{devlin2018bert},
so that ZEN can be compared with BERT on a fair basis.
We use the same parameter setting for the n-gram encoder as in BERT,
except that we only use 6 layers and set 128 as the max length of n-grams\footnote{That is, we extract up to 128 n-grams per instance.}. The resulting ZEN requires only 20\% additional inference time (averaged by testing on the seven tasks) over the original BERT base model.
We adopt mixed precision training \citep{micikevicius2017mixed} by the Apex library\footnote{https://github.com/NVIDIA/apex} to speed up the training process.
Each ZEN model is trained simultaneously on 4 NVIDIA Tesla V100 GPUs with 16GB memory.

Our task-specific fine-tuning uses similar hyper-parameters reported in \citet{cui2019pre},
with slightly different settings on max input sequence length and batch size
for better utilization of computational resources.
Specifically, we set max length to 256 for CWS and POS, and 96 for their batch size.
For NER, SPM and NLI, we set both the max length and batch size to 128.
For the other two tasks, DC and SA, we set the max length and batch size to 512 and 32, respectively.

\begin{table*}[t]
\begin{center}
\small
\begin{sc}
\begin{tabular}{p{1.5cm}R{0.73cm}R{0.73cm}R{0.73cm}R{0.73cm}R{0.73cm}R{0.73cm}R{0.73cm}R{0.73cm}R{0.73cm}R{0.73cm}R{0.73cm}R{0.73cm}}
\toprule
 & 
CWS& \multicolumn{2}{c}{POS} & NER & \multicolumn{2}{c}{DC} & \multicolumn{2}{c}{SA} & \multicolumn{2}{c}{SPM} & \multicolumn{2}{c}{NLI} \\ 
\cline{2-13}
& Test & Dev & Test & Test & Dev & Test & Dev & Test & Dev & Test & Dev & Test   \\ 
\midrule
BERT (R)~~       &95.14 & 93.64 &93.23  & 87.11 & 96.02  & 95.77 & 93.41 & 92.33 & 85.62 & 85.53 & 72.12 & 71.44  \\
ZEN (R)    & \textbf{96.05} & \textbf{93.79} & \textbf{93.37}  & \textbf{88.39} & \textbf{96.11} & \textbf{96.05} & \textbf{93.92} & \textbf{93.51} & \textbf{86.12} & \textbf{85.78} & \textbf{72.66} & \textbf{72.31}  \\
\bottomrule
\end{tabular}
\end{sc}
                \vspace{-0.3cm}
\end{center}
\caption{The performance of BERT and ZEN on seven NLP tasks when they are trained on a small corpus.
}
\label{tab:CompareOnSmallCorpus}
\vskip -0.5em
\end{table*}

\section{Experimental Results}

\subsection{Overall Performance}



The first experiment is to compare ZEN and BERT with respect to their performance on the aforementioned NLP tasks.
In this experiment, ZEN and BERT use two settings, i.e., training from (R): randomly initialized parameters and (P): pre-trained model, which is the Google released Chinese BERT base model.
The results are reported in Table \ref{tab:OverallPerformance}, with the evaluation metrics for each task denoted in the second row.
Overall, in both R and P settings, ZEN outperforms BERT in all seven tasks, which clearly indicates the advantage of introducing n-grams into the encoding of character sequences.
This observation is similar to that from \citet{dos2014deep, lample2016neural, bojanowski2017enriching, liu2019encoding}.
%
In detail, when compare R and P settings, the performance gap between ZEN (P) and BERT (P) is larger than that in their R setting,
which illustrates that learning an encoder with reliable initialization is more important and integrating n-gram information contributes a better enhancement on well-learned encoders.
For two types of tasks, it is noticed that token-level tasks, i.e., CWS, POS and NER, demonstrate
a bigger improvement of ZEN over BERT than that of sentence-level tasks.
where the potential boundary information presented by n-grams are essential to provide a better guidance to label each character.
Particularly for CWS and NER, these boundary information are directly related to the outputs.
Similarly, sequence-level tasks show a roughly same trend on the improvement of ZEN over BERT,
which also shows the capability of combining both character and n-gram information in a text encoder.
The reason behind this improvement is that in token-level tasks,
where high-frequent n-grams\footnote{Such as fixed expressions and common phrases, which may have less varied meanings than other ordinary combinations of characters and random character sequences.} in many cases are valid chunks in a sentence that carry key semantic information.

We also compare ZEN (P) and existing pre-trained encoders on the aforementioned NLP tasks, with their results listed in the middle part of Table \ref{tab:OverallPerformance}.\footnote{We only report the performance on their conducted tasks.}
Such encoders include BERT-wwm \citep{cui2019pre}, ERNIE 1.0 \citep{sun2019ernie}, ERNIE 2.0 (B) \citep{sun2019ernie2}, ERNIE 2.0 (L) \citep{sun2019ernie2}, NEZHA (B) and (L) \citep{wei2019nezha} where B and L denote the base and large model of BERT, respectively. Note that although there are other pre-trained encoders with exploiting entity knowledge or multi-model signals, they are not compared in this paper because external information are required in their work.
In fact, even though without using such external information, ZEN still achieves the state-of-the-art performance on many of the tasks experimented.

%
In general, the results clearly indicate the effectiveness of ZEN.
In detail, for the comparison between ZEN and BERT-wwm, it shows that, when starting from pre-trained BERT, ZEN outperforms BERT-wwm on all tasks that BERT-wwm has results reported.
This observation suggests that explicitly representing n-grams and integrating them into BERT has its advantage over using masking strategy,
and using n-grams rather than word may have better tolerance on error propagation since word segmentation is unreliable in many cases.
%
The comparison between ZEN and ERNIE encoders also illustrates the superiority of enhancing BERT with n-grams.
For example, ZEN shows a consistent improvement over ERNIE 1.0 even though significantly larger non-public datasets were utilized in their pre-training.
Compared to ERNIE 2.0, which used many more pre-training tasks and significantly more non-public training data, 
ZEN is still competitive on SA, SPM and NLI tasks.
Particularly,
ZEN outperforms ERNIE 2.0 (B) on SA (\textsc{Test}) and SPM (\textsc{Test}), which indicates that n-gram enhanced character-based encoders of ZEN can achieve performance comparable to approaches using significantly more resources.
Since the two approaches are complementary to each other, one might be able to combine them to achieve higher performance. 
%
Moreover, ZEN and ERNIE 2.0 (L) have comparable performance on some certain tasks (e.g., SA and SPM), which further confirms the power of ZEN even though the model of ERNIE 2.0 is significantly larger. 
Similar observation is also drawn from the comparison between ZEN and NEZHA, where ZEN illustrates its effectiveness again when compared to a model that learning with larger model and more data, as well as more tricks applied in pre-training.
%
However,
for NLI task, ZEN's performance is not as good as ERNIE 2.0 and NEZHA (B \& L), which further indicates that their model are good at inference task owing to their larger model setting and large-scale corpora have much more prior knowledge.
%

\subsection{Pre-training with Small Corpus}
Pre-trained models usually require a large corpus to perform its training. However, in many applications in specialized domains, a large corpus may not be available. For such applications with limited training data, ZEN, with n-gram enhancement, is expected to encode text much more effectively.
Therefore,
to further illustrate the advantage of ZEN,
we conduct an experiment that uses a small corpus to pre-train BERT and ZEN.
In detail, we prepare a corpus with $1/10$ size of the entire Chinese Wikipedia 
by randomly selecting sentences from it.
Then all encoders are pre-trained on it with random initialization and tested on the same NLP tasks in the previous experiment.
The results are reported in Table \ref{tab:CompareOnSmallCorpus}.
%
In general, same trend is shown in this experiment when compared with that in the previous one,
where ZEN constantly outperform BERT in all task.
This observation confirms that representing n-grams provides stable enhancement when our model is trained on corpora with different sizes.
In detail, these results also reveals that n-gram information helps more on some tasks, e.g., CWS, NER, NLI, over the others.
The reason is not surprising since that boundary information carried by n-grams can play a pivotal role in these tasks.
Overall,
this experiment simulates the situation of pretraining a text encoder with limited data, which could be a decisive barrier in pre-training a text encoder in the cold-start scenario,
and thus demonstrates that ZEN has its potential to perform well in this situation.


\section{Analyses}
We analyze ZEN
with several factors affecting its performance.
Details are illustrated in this section.

\begin{figure}[t]
	\begin{center}
		\includegraphics[scale=0.21, trim=15 20 0 10]{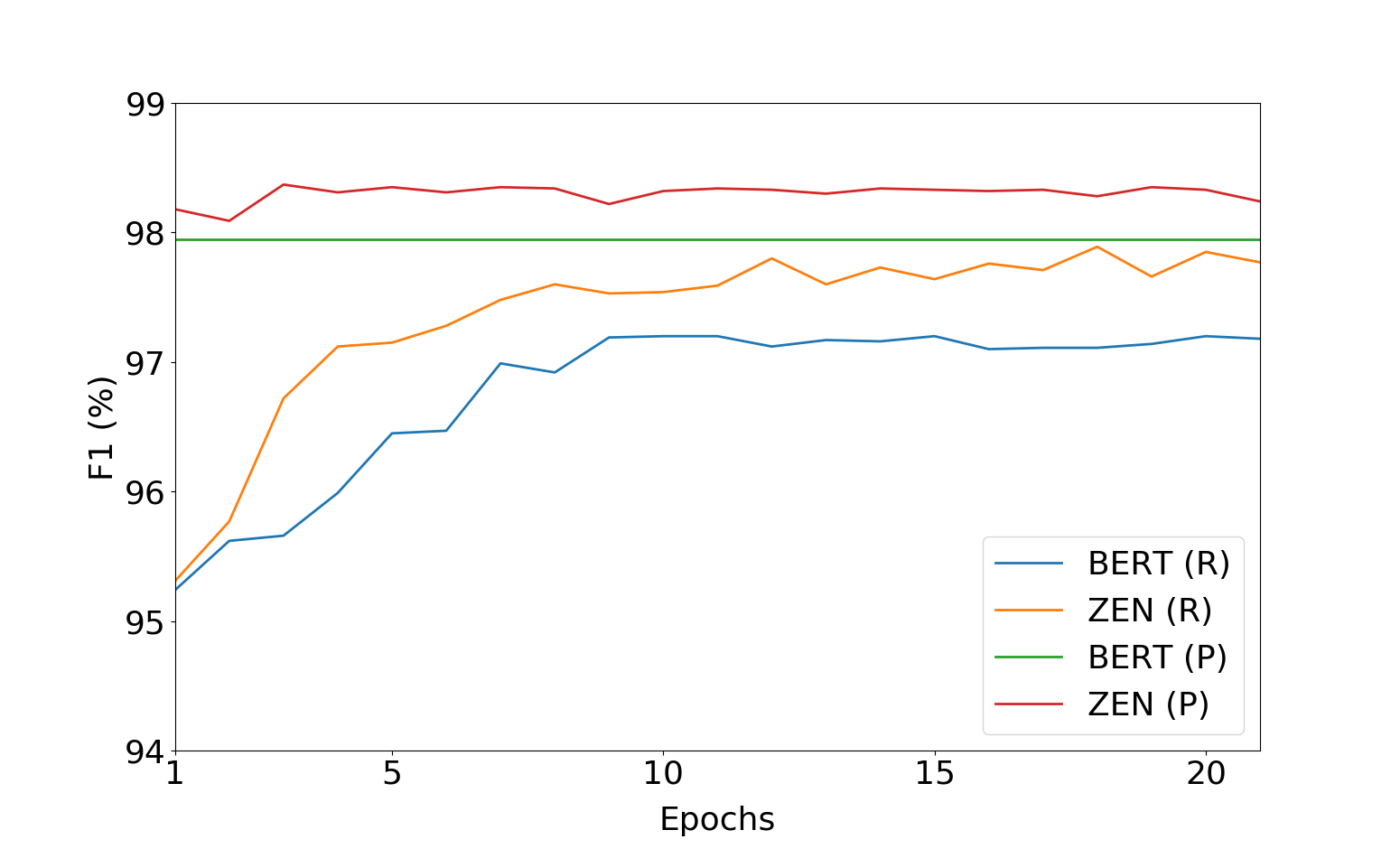}
        \caption{CWS performance against training epochs of BERT and ZEN with different parameter initialization.}
        \label{CWS performance against Training Epochs}
	\end{center}
	\vspace{-2.0em}
\end{figure}

\begin{figure}[t]
	\begin{center}
		\includegraphics[scale=0.21, trim=15 20 0 0]{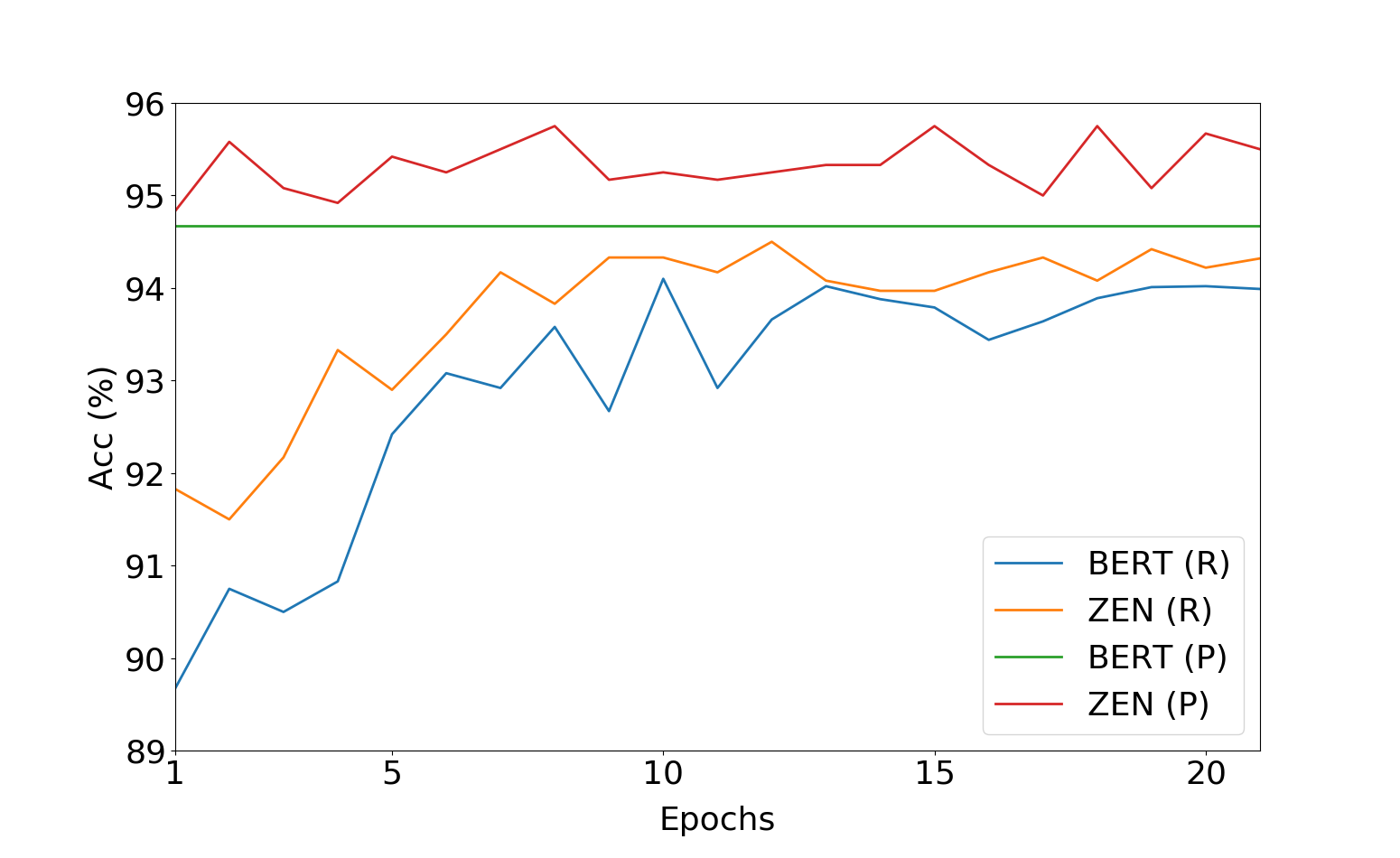}
        \caption{SA performance against training epochs of BERT and ZEN with different parameter initialization.}
        \label{SA performance against Training Epochs}
	\end{center}
	\vspace{-1.0em}
\end{figure}

%
%

%
%

\subsection{Effects of Pre-training Epochs}

The number of pretraining epochs is another factor affecting the performance of pre-trained encoders.
In this analysis,
we use CWS and SA as two probing tasks to test the performance of different encoders (BERT and ZEN) against the number of pretraining epochs.
The pretrained models at certain epochs are fine-tuned on these tasks, and the
results are illustrated in Figure \ref{CWS performance against Training Epochs} and \ref{SA performance against Training Epochs}. We have the following observations.
First, for both P and R models, ZEN shows  better curves than those of BERT in both tasks, which indicates that ZEN achieves higher performance at comparable pretraining stages.
Second, for R settings, ZEN shows a noticeable faster convergence than BERT, especially during the first few epochs of pretraining. This demonstrates that n-gram information improves the encoder's performance when pretraining starts from random initialization.

\subsection{Effects of N-gram Extraction Threshold}

\begin{figure}[t]
	\begin{center}
		\includegraphics[scale=0.21, trim=15 20 0 10]{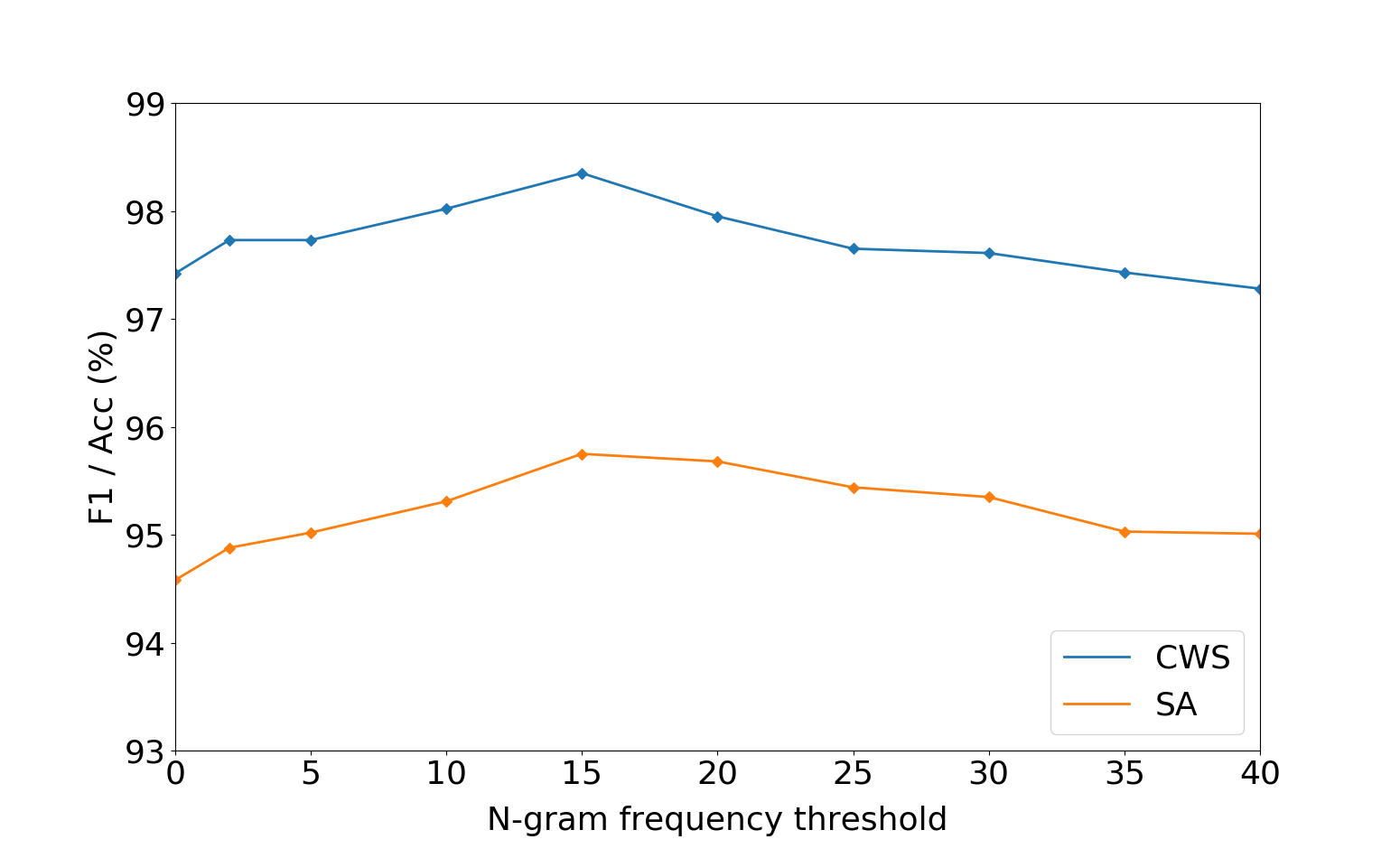}
        \caption{CWS and SA performance of ZEN against frequency threshold of constructing n-gram lexicons.}
        \label{fig:n-gram frequency threshold}
	\end{center}
	\vspace{-2.0em}
\end{figure}

\begin{figure}[t]
	\begin{center}
		\includegraphics[scale=0.21, trim=15 20 0 0]{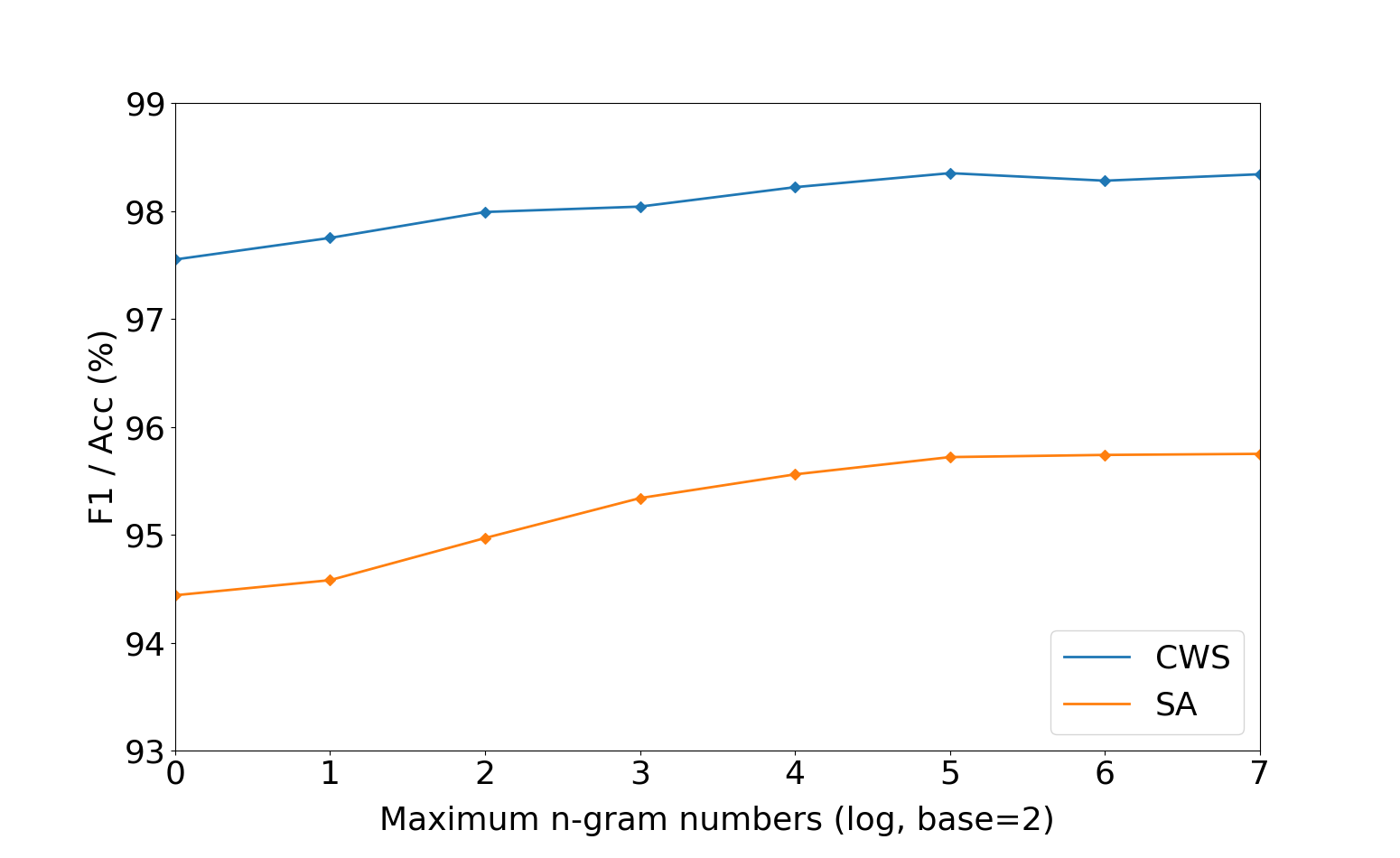}
        \caption{CWS and SA performance of ZEN against maximum n-gram numbers for training each instance.}
        \label{fig:n-gram proportion}
	\end{center}
	\vspace{-1.0em}
\end{figure}

To explore how n-gram extraction cutoff threshold affects the performance of ZEN,
we test it with different thresholds for n-gram lexicon extraction.
Similar to the previous experiment, we also use CWS and SA as the probe tasks in this analysis.
%

The first analysis on threshold-performance relations is demonstrated in Figure \ref{fig:n-gram frequency threshold}, where we set the threshold ranging from 0 to 40 and use the max number of 128 n-grams in pre-training.
In doing so, we observe that 
the best performed ZEN on both tasks is obtained when the threshold is set to 15, where increasing the threshold value under 15 causes improved performance of ZEN and vice versa when it gets over 15.
This observation confirms that either too many (lower threshold) or too few (higher threshold) n-grams in the lexicon are less helpful in enhancing ZEN's performance, since there exists a balance between introducing enough knowledge and noise.
%
%
%

For the second analysis,
when an optimal threshold is given (i.e., 15), one wants to know the performance of ZEN with different maximum number of n-grams in pre-training for each input sequence.
In this analysis we test such number ranging from $0$ (no n-grams encoded in ZEN) to $128$, with
the results shown in Figure \ref{fig:n-gram proportion} (X-axis is in log view with base 2).
It shows that the number $32$ ($2^5$) gives a good tradeoff between performance and computation, although there is a small gain by using more n-grams.
This analysis illustrates that ZEN only requires a small numbers of n-grams to achieve good performance.

\begin{figure}[t]
	\begin{center}
		\includegraphics[scale=0.32, trim=10 20 0 0]{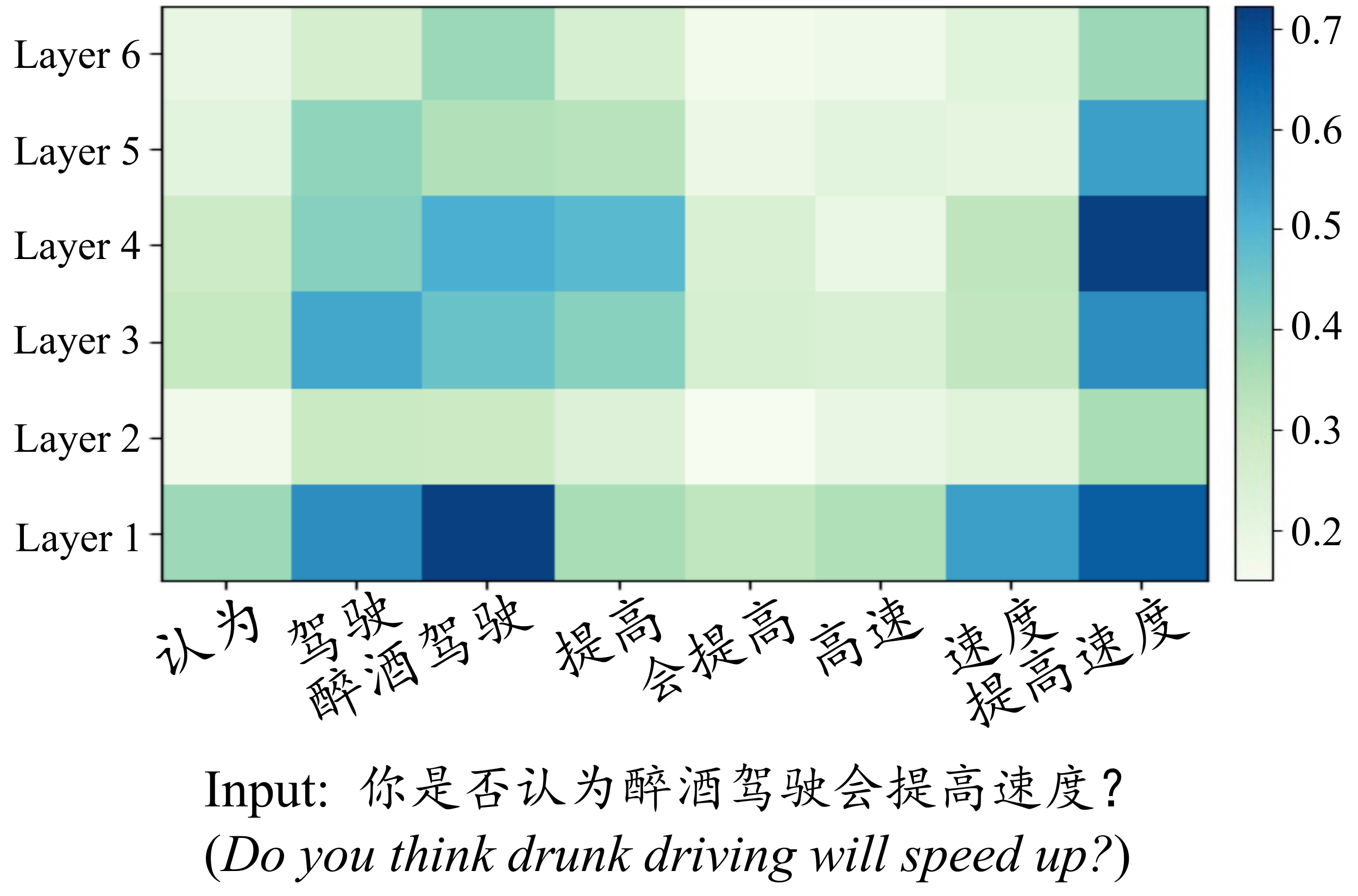}
        \caption{The heatmap of n-grams encoded by ZEN across different layers for an example sentence.}
        \label{fig:visualization-1}
	\end{center}
	\vspace{-1.0em}
\end{figure}

\subsection{Visualization of N-gram Representations}

\begin{figure}[t]
	\begin{center}
		\includegraphics[scale=0.32, trim=10 20 0 0]{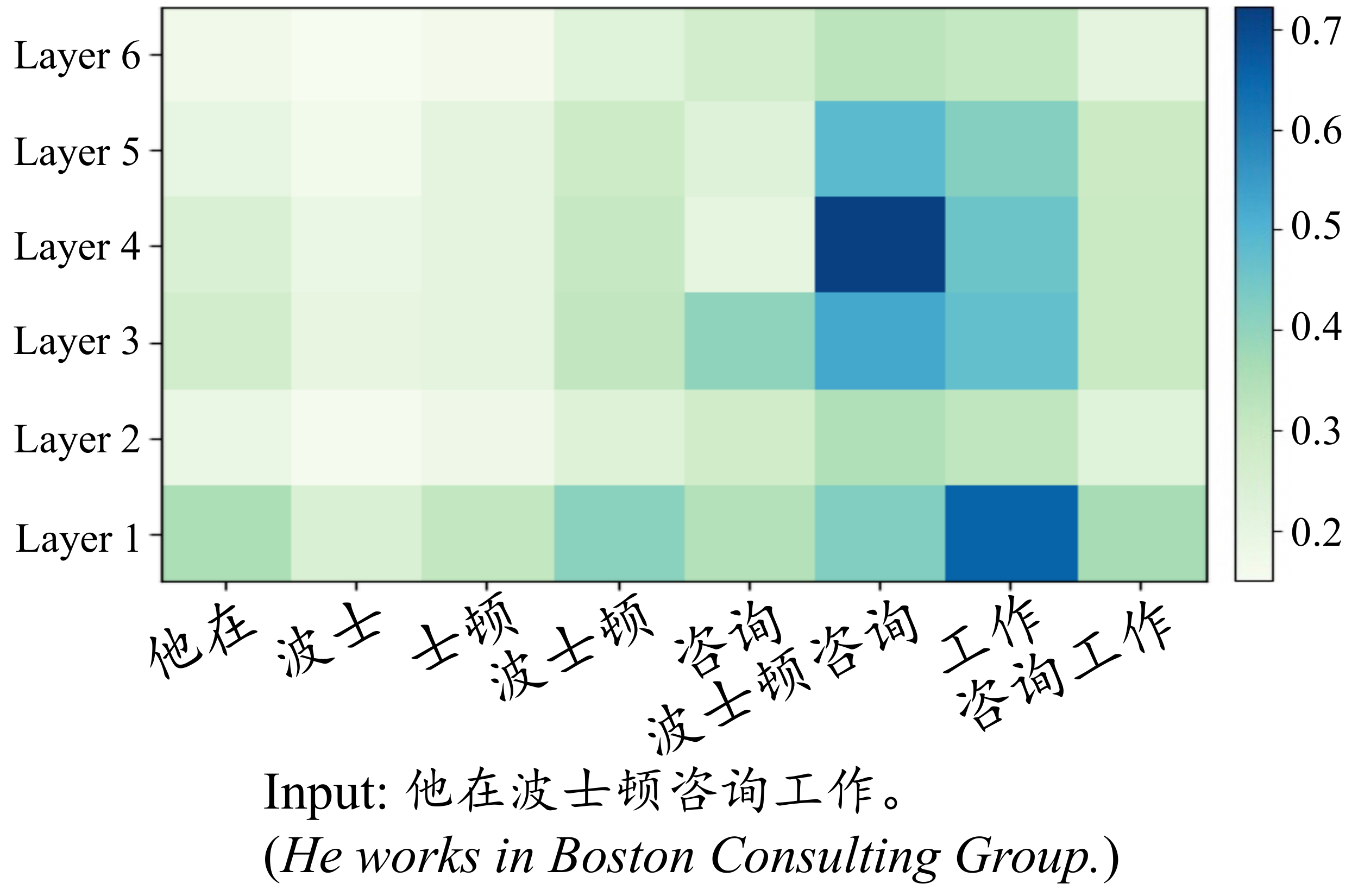}
        \caption{The heatmap of n-grams encoded by ZEN across different layers for an example sentence.}
        \label{fig:visualization-2}
	\end{center}
	\vspace{-1.0em}
\end{figure}

In addition to quantitative analysis,
we also conduct case studies on some certain instances to further illustrate the effectiveness of n-gram representations in pre-training ZEN.
Figure \ref{fig:visualization-1} and \ref{fig:visualization-2} visualize the weights of extracted n-grams from two input instances when they are encoded by ZEN across different layers.
In general, `valid' n-grams are more favored than others, e.g., \begin{CJK*}{UTF8}{gkai}提高\end{CJK*} (\textit{improve}), \begin{CJK*}{UTF8}{gkai}波士顿\end{CJK*} (\textit{Boston}) have higher weights than \begin{CJK*}{UTF8}{gkai}会提高\end{CJK*} (\textit{will improve}) and \begin{CJK*}{UTF8}{gkai}士顿\end{CJK*} (\textit{Ston}),
especially those ones that have cross ambiguities in the context, e.g., \begin{CJK*}{UTF8}{gkai}高速\end{CJK*} (\textit{high speed}) should not be considered in the first instance so that \begin{CJK*}{UTF8}{gkai}速度\end{CJK*} (\textit{speed}) has a higher weight than it.
This observation illustrates that ZEN is able to not only distinguish those phrasal n-grams to others but also select appropriate ones according to the context.
Interestingly, for different layers, long (and valid) n-grams, e.g., \begin{CJK*}{UTF8}{gkai}提高速度\end{CJK*} (\textit{speed up}) and \begin{CJK*}{UTF8}{gkai}波士顿咨询\end{CJK*} (\textit{Boston consulting group}),
tend to receive more intensive weights at higher layers,
which implicitly indicates that such n-grams contain more semantic rather than morphological information. We note that information encoded in BERT follows a similar layer-wise order
as what is suggested in \citet{jawahar2019does}.
The observations from this case study, as well as the overall performance in previous experiments, suggest that integration of n-grams in ZEN not only enhances the representation power of character-based encoders, but also provides a potential solution to some text analyzing tasks, e.g., chunking and keyphrase extraction.






\section{Related Work}

Representation learning of text attracts much attention in recent years, with the rise of deep learning in NLP \citep{collobert2011-JMLR, mikolov2013efficient, pennington2014glove}.
There are considerable interests in representing text with contextualized information \cite{ling2015-NAACL-HLT, melamud-etal-2016-context2vec, bojanowski2017enriching, song-etal-2017-learning, song-etal-2018-directional, peters-etal-2018-deep, ijcai2018-607}.
Following this paradigm,  pre-trained models have been proposed and are proven useful in many NLP tasks \citep{devlin2018bert, radford2018improving, radford2019language, yang2019xlnet}. 
%
%
%
In detail, such models can be categorized into two types: autoregressive and autoencoding encoders.
The former models behave like normal language models that predict the probability distributions of text units following observed texts.
These models, such as 
GPT \citep{radford2018improving} and GPT2 \citep{radford2019language}, are trained to encode a uni-directional context.
Differently, the autoencoding models, such as
BERT \citep{devlin2018bert} and XLNet \citep{yang2019xlnet}, leverage bidirectional context, and encode text by reconstructing the masked tokens in each text instance according to their context from both sides.

Because words carry important linguistic information in Chinese, many enhanced pre-train models are proposed specifically for Chinese that can utilize word-level information in one way or another. 
For example, ERNIE 1.0 \citep{sun2019ernie} adopted a multi-level masking strategy performed on different level of texts;
its improved version, ERNIE 2.0 \citep{sun2019ernie2} used continual pre-training strategy which is benefited from multi-task learning with more parameters in the model.
Recently, BERT-wwm \citep{cui2019pre} enhanced Chinese BERT with a simple masking of whole-words. 
In addition, there are other recent studies that enhanced BERT for Chinese language processing, such as optimizing training via special optimization techniques \cite{wei2019nezha} or from prior knowledge \cite{liu2019k}.
%
%
%
%
%
All the studies revealed that processing on larger granularity of text is helpful in Chinese, which is consistent with previous findings in many Chinese NLP tasks \citep{song-etal-2009-transliteration, song-xia-2012-using, wu2015named, chang1802hybrid, higashiyama-etal-2019-incorporating}.
However, previous approach are limited to the use of weak supervision, i.e., masking, to incorporate word/phrase information.
ZEN thus provides an alternative solution that explicitly encodes n-grams into character-based encoding, 
which is effective for downstream NLP tasks.

\section{Conclusion}
In this paper, we proposed ZEN, a pre-trained Chinese text encoder enhanced by n-gram representations,
where different combinations of characters are extracted, encoded and integrated in training a backbone model, i.e., BERT.
In ZEN, given a sequence of Chinese characters, n-grams are extracted and their information are effectively incorporated into the character encoder.
Different from previous work, ZEN provides an alternative way of learning larger granular text for pre-trained models, where the structure of BERT is extended by another Transformer-style encoder to represent the extracted n-grams for each input text instance.
%

Experiments on several NLP tasks demonstrated the validity and effectiveness of ZEN.
Particularly, state-of-the-art results were obtained while ZEN only uses BERT base model requiring less training data and no knowledge from external sources compared to other existing Chinese text encoders.
%
Further analyses of ZEN are conducted,
showing that ZEN is efficient and able to learn with limited data.
We note that ZEN employs a different method to incorporate word information that is complementary to some other previous approaches.
Therefore it is potentially beneficial to combine it with previous approaches suggested by other researchers, as well as to other languages.
%

\bibliography{acl2019}
\bibliographystyle{acl_natbib}
\end{CJK}
\end{document}